\documentclass{article}
\usepackage{mystylefile}
\usepackage{subfigure}
\usepackage{times}
\usepackage{color}
\usepackage{multirow}
\usepackage{algpseudocode}
\usepackage{bm}
\usepackage{xr}
\usepackage{natbib}

\usepackage[preprint]{nips_2018}

\usepackage[utf8]{inputenc} 
\usepackage[T1]{fontenc} 
\usepackage{hyperref} 
\usepackage{url} 
\usepackage{booktabs} 
\usepackage{amsfonts} 
\usepackage{nicefrac} 
\usepackage{microtype} 
\usepackage{float}

\newcommand{\dx}{d_{\mathrm{x}}}

\newcommand{\dy}{d_{\mathrm{y}}}

\newcommand{\calX}{\mathcal{X}}
\newcommand{\calY}{\mathcal{Y}}

\newcommand{\calC}{\mathcal{C}}
\newcommand{\calF}{\mathcal{F}}

\newcommand{\calD}{\mathcal{D}}
\newcommand{\calFx}{\mathcal{F}_{\mathrm{x}}}

\newcommand{\calFy}{\mathcal{F}_{\mathrm{y}}}

\def\<{\langle}
\def\>{\rangle}
\begin{document}
\title{Neural-Kernelized Conditional Density Estimation}
\author{Hiroaki Sasaki$^{1,2}$ \\ \texttt{hsasaki@is.naist.jp} \And
  Aapo Hyv{\"a}rinen$^{2,3}$ \\ \texttt{aapo.hyvarinen@helsinki.fi}
  \AND
  \vspace{-8mm} \\ $^1$Div. of Info. Sci.\\ NAIST, Japan \And \vspace{-8mm} \\
  $^2$Gatsby Unit\\ UCL, UK\\ \And \vspace{-8mm} \\ $^3$ Dept. of
  Comp. Sci.\\ Univ. Helsinki, Finland }


%
\maketitle
\begin{abstract}
  Conditional density estimation is a general framework for solving
  various problems in machine learning. Among existing methods,
  non-parametric and/or kernel-based methods are often difficult to
  use on large datasets, while methods based on neural networks
  usually make restrictive parametric assumptions on the probability
  densities. Here, we propose a novel method for estimating the
  conditional density based on score matching. In contrast to existing
  methods, we employ scalable neural networks, but do not make
  explicit parametric assumptions on densities. The key challenge in
  applying score matching to neural networks is computation of the
  first- and second-order derivatives of a model for the
  log-density. We tackle this challenge by developing a new
  neural-kernelized approach, which can be applied on large datasets
  with stochastic gradient descent, while the reproducing kernels
  allow for easy computation of the derivatives needed in score
  matching. We show that the neural-kernelized function approximator
  has universal approximation capability and that our method is
  consistent in conditional density estimation. We numerically
  demonstrate that our method is useful in high-dimensional
  conditional density estimation, and compares favourably with
  existing methods. Finally, we prove that the proposed method has
  interesting connections to two probabilistically principled
  frameworks of representation learning: Nonlinear sufficient
  dimension reduction and nonlinear independent component analysis.
\end{abstract}
 \section{Introduction}
 Given observations of two random variables $\bm{y}$ and $\bm{x}$,
 estimating the conditional probability density function of $\bm{y}$
 given $\bm{x}$ is a general framework for solving various problems in
 machine learning. In particular, when the conditional density
 $p(\bm{y}|\bm{x})$ is multimodal, the conventional prediction by the
 conditional mean of $\bm{y}$ given $\bm{x}$ is rather useless, and
 conditional density estimation becomes very
 useful~\citep{carreira2000reconstruction,einbeck2006modelling,chen2016nonparametric}.
 Also, in regression problems where noise is heavy-tailed or
 asymmetric, the conditional mean is not necessarily a suitable
 quantity for prediction, and again, one may want to estimate the
 conditional density
 $p(\bm{y}|\bm{x})$~\citep{yao2012local,feng2017statistical}.

 However, the importance of conditional density estimation is not
 limited to supervised learning. A well-known example is structure
 estimation in graphical models~\citep{pattern2006bishop}.
 Furthermore, conditional density estimation seems to be closely
 related to \emph{representation learning} (or feature learning),
 which has recently received a great deal of attention due to the
 success of deep neural networks~\citep{bengio2013representation}. For
 example, \emph{sufficient dimension reduction} (SDR) is a rigorous
 framework for supervised dimension reduction based on the conditional
 density, where the goal is to estimate a lower-dimensional
 representation of $\bm{x}$ without losing information to predict
 $\bm{y}$~\citep{li1991sliced,cook1998regression}.  In addition,
 \emph{nonlinear independent component analysis} (ICA) provides a
 principled way for representation learning in unsupervised learning,
 which is aimed at recovering latent nonlinear sources (or
 representation) that generated the input data
 $\bm{x}$. \citet{hyvarinen2016unsupervised} recently proposed a
 nonlinear ICA method with a rigorous proof of recovering the sources
 (i.e., identifiability) where the conditional probability density of
 $\bm{x}$ given time-segment labels is estimated via logistic
 regression. Thus, we see that conditional density estimation is
 closely related to representation learning as well.  In this paper,
 we restrict ourselves to the case where $\bm{y}\in\R{\dy}$ and
 $\bm{x}\in\R{\dx}$ are both continuous-valued.

 In a nonparametric setting, a number of methods for conditional
 density estimation have been proposed.  \citet{fan996estimation}
 estimates conditional densities based on a local polynomial
 expansion. However, it is known that the local polynomial expansion
 does not scale well with the dimensionality of
 data. \citet{sugiyama2010least} proposed another method called the
 \emph{least-squares conditional density estimation} (LSCDE), which
 directly fits a kernel model to the true conditional density ratio
 under the squared loss. Another method with reproducing kernels has
 been proposed recently based on the Fisher divergence, estimating
 conditional densities up to their partition
 functions~\citep{pmlr-v84-arbel18a}. A drawback is that since these
 methods rely on reproducing kernels, it is not easy to apply them to
 large datasets. The Nystr{\"o}m approximation~\citep{rudi2015less},
 where only a subset of data samples is used as the centers of kernel
 functions, is a remedy to make kernel methods more scalable, but it
 may not allow us to accurately estimate highly complex
 functions~\citep{bengio2006curse}, especially with high-dimensional
 data.
 
 To accurately estimate conditional densities, we need models which
 have high expressive power, yet are scalable to a large amount of
 data---such as deep neural networks. However, since such a model is
 often very complex, it may be almost impossible to compute the
 partition function of the conditional density, which is the
 fundamental challenge in learning deep neural
 networks~\citep[Chapter~18]{goodfellow2016deep}. Presumably due to
 this reason, some previous work has combined neural networks with a
 parametric density model: The \emph{mixture density network}
 (MDN)~\citep{bishop1994mixture} employs a mixture model for the
 conditional density, and each parameter in the mixture density model
 is estimated as a feedforward neural network.  Another approach is
 based on variational
 methods~\citep{tang2013learning,sohn2015learning}, which also makes
 parametric assumptions on the densities. Although these methods may
 work rather well in some applications, the parametric assumption on
 the densities seems too restrictive to us.
 
 In this paper, we propose a novel method of estimating a conditional
 density function up to the partition function. In stark contrast to
 previous work, we use neural networks but do not make explicit
 parametric assumptions on the probability densities. Our approach is
 to employ the Fisher divergence, and more precisely its computation
 by score
 matching~\citep{cox1985penalty,hyvarinen2005estimation,SasakiHS14clustering,JMLR:v18:16-011},
 for conditional density estimation. However, the significant
 challenge is that score matching requires us to compute the first-
 and second-order derivatives of the model with respect to each
 coordinate in $\bm{y}$, and thus it is difficult to apply to learning
 neural networks~\citep{martens2012estimating}. Here, we tackle this
 challenge by combining the benefits of the neural network approach
 with those of reproducing kernel Hilbert spaces (RKHS). Basically, we
 develop a neural-kernelized approach in which the dependence of the
 conditional density on $\bm{y}$ is modelled in an RKHS and the
 dependence on $\bm{x}$ by a neural network. This means the
 derivatives only need to be computed in the RKHS, which is
 computationally easy, while the neural network part still allows for
 powerful deep representation learning to happen in high-dimensional
 spaces and with big data sets. Since $\bm{y}$ usually has much lower
 dimension than $\bm{x}$ as seen especially in supervised learning
 problems, our method is efficient in many applications of conditional
 density estimation.
 
 We prove that our function approximator following the
 neural-kernelized approach has the universal approximation capability
 under certain conditions, and thus potentially approximates a wide
 range of continuous functions of $\bm{y}$ and $\bm{x}$. We further
 prove that our method for conditional density estimation is
 consistent. In artificial and benchmark data comparisons based on
 conditional likelihood, our method is demonstrated to be useful in
 high-dimensional conditional density estimation, and compares
 favourably with competing methods.  Finally, we explore very
 interesting connections of our method to representation learning. It
 turns out that under some conditions, our method for conditional
 density estimation automatically provides novel methods for two
 different principled frameworks related to representation learning:
 nonlinear SDR and nonlinear ICA. More specifically, the neural
 network in our model provides a useful lower-dimensional
 representation as in SDR, while it recovers the latent independent
 sources as in ICA.
 \section{Neural-Kernelized Conditional Density Estimator}
 We start by formulating the problem of conditional density
 estimation. Then, we propose a simple function approximation
 framework, and validate it using the well-known concept of universal
 approximation capability. Next, a statistical divergence is proposed
 based on score matching. Finally, a practical learning algorithm is
 developed by combining neural networks with reproducing kernels,
 together with the score matching divergence.
 \paragraph{Problem Formulation:}
 Suppose that we are given $T$ pairs of data samples drawn from the
 joint distribution with density $p(\bm{y},\bm{x})$:
 \begin{align}
   \calD:=\left\{(\bm{y}(t)^{\top},\bm{x}(t)^{\top})^{\top}~|~%
     \bm{y}(t)=(y_{1}(t),\dots,y_{\dy}(t))^{\top},
     \bm{x}(t)=(x_{1}(t),\dots,x_{\dx}(t))^{\top}\right\}_{t=1}^T,
   \label{dataset}
 \end{align}
 where $\bm{x}(t)$ and $\bm{y}(t)$ denote the $t$-th observations of
 $\bm{x}$ and $\bm{y}$, respectively, and $\dx$ and $\dy$ denote the
 dimensions of $\bm{x}$ and $\bm{y}$, respectively. For example, in
 supervised learning, $\bm{y}$ would be the output data. We will also
 consider the case of unsupervised learning, where following
 \citet{hyvarinen2016unsupervised}, $\bm{y}$ could a time index (more
 on this below). Our primary interest is to estimate the logarithmic
 conditional density of $\bm{y}$ given $\bm{x}$ up to the partition
 function from $\calD$:
 \begin{align}
   \log p(\bm{y}|\bm{x}) = \log q(\bm{y}|\bm{x}) -\log Z(\bm{x}),
   \label{log-cond}
 \end{align}
 where $q$ is an unnnormalized conditional density, and $Z(\bm{x})$ is
 the partition function.
 
 \paragraph{Function Approximation Framework:}
 In this paper, we propose a function approximation framework where we
 estimate $\log q(\bm{y}|\bm{x})$ under the following form:
 \begin{align}
   \log q(\bm{y}|\bm{x}) =
   \bm{w}(\bm{y})^{\top}\bm{h}(\bm{x})=\sum_{i=1}^dw_i(\bm{y})h_i(\bm{x}),
   \label{model-form}
 \end{align}
 where $\bm{w}(\bm{y})=(w_1(\bm{y}),\dots,w_d(\bm{y}))^{\top}$ and
 $\bm{h}(\bm{x})=(h_1(\bm{x}),\dots,h_{d}(\bm{x}))^{\top}$ are
 vector-valued functions. 

 The model~\eqref{model-form} in our framework may seem to be of a
 restricted form, and unable to express general continuous functions
 of $\bm{y}$ and $\bm{x}$. However, we next prove that the model can
 actually approximate a wide range of continuous functions. To this
 end, we employ the well-known concept of \emph{universal
   approximation capability}:
  \begin{definition}[Universal approximation capability]
    Let $\calC(\calX)$ be the set of all continuous functions on a
    domain $\calX$. Then, a set of functions on $\calX$,
    $\calF(\calX)$ has the universal approximation capability to
    $\calC(\calX)$ if $\calF(\calX)$ is a dense subset of
    $\calC(\calX)$, i.e., for all function $g\in\calC(\calX)$ and
    $\epsilon>0$, there always exists $f\in\calF(\calX)$ such that
    $\sup_{\bm{x}\in\calX}|f(\bm{x})-g(\bm{x})|\leq\epsilon$.
  \end{definition}
  
  Let us express two sets of functions on domains $\bm{y}\in\calY$ and
  $\bm{x}\in\calX$ by $w_i\in\calFy(\calY)$ and $h_i\in\calFx(\calX)$
  for all $i$, respectively.  Eq.\eqref{model-form} indicates that
  $\log q(\bm{y}|\bm{x})$ is estimated as an element in the following
  set of functions:
  \begin{align*}
   \calFy(\calY)\odot\calFx(\calX)
   :=\left\{w\cdot h~|~w\in\calFy(\calY), h\in\calFx(\calX)\right\}.
  \end{align*}
  $\calFy(\calY)\odot\calFx(\calX)$ does not necessarily have the
  universal approximation capability to $\calC(\calY\times\calX)$ in
  general. Thus, it is important to understand under what conditions
  $\calFy(\calY)\odot\calFx(\calX)$ has the universal approximation
  capability. The following proposition states sufficient conditions.
  \begin{proposition}
    \label{prop:universal-approximation}
    Assume that (i) both $\calY$ and $\calX$ are compact, (ii) every
    function in $\calFy(\calY)$ and $\calFx(\calX)$ is continuous, and
    (iii) both $\calFy(\calY)$ and $\calFx(\calX)$ have the universal
    approximation capability to $\calC(\calY)$ and $\calC(\calX)$,
    respectively. Then, $\calFy(\calY)\odot\calFx(\calX)$ has the
    universal approximation capability to $\calC(\calY\times\calX)$,
    i.e., $\calFy(\calY)\odot\calFx(\calX)$ is a dense subset of
    $\calC(\calY\times\calX)$.
  \end{proposition}
  We omit the proof because it is essentially the same
  as~\citet[Thereom~II.I]{waegeman2012kernel}.

  Proposition~\ref{prop:universal-approximation} implies that the
  model~\eqref{model-form} can approximate a wide range of continuous
  functions on $\calY\times\calX$ when both $\calFy(\calY)$ and
  $\calFx(\calX)$ have the universal approximation capability. Thus,
  we should use function approximators for $w_i$ and $h_i$ such that
  both of them have the universal approximation capability.
  
  It is well-known that the function set of feedforward neural
  networks have the universal approximation
  capability~\citep{hornik1991approximation}. Another example of
  function sets with the universal approximation capability is
  reproducing kernel Hilbert spaces (RKHSs) with universal kernels
  whose example is the Gaussian
  kernel~\citep{micchelli2006universal}. Motivated by these facts and
  Proposition~\ref{prop:universal-approximation}, we employ both
  neural networks and RKHSs in the learning algorithm below.
  \paragraph{Statistical Divergence based on Score Matching}
  To estimate $\log q(\bm{y}|\bm{x})$, we employ \emph{score
    matching}~\citep{hyvarinen2005estimation}. It is based on fitting
  a model $g(\bm{y},\bm{x})$ to $\log q(\bm{y}|\bm{x})$ under the
  \emph{Fisher
    divergence}~\citep{cox1985penalty,SasakiHS14clustering,JMLR:v18:16-011},
  defined as
  \begin{align*}
    J(g):=\frac{1}{2}\iint
    \|\nabla_{\bm{\bm{y}}}g(\bm{y},\bm{x})-\nabla_{\bm{y}}\log
    q(\bm{y}|\bm{x})\|^2 p(\bm{y},\bm{x}) \intd\bm{y}\intd\bm{x},
  \end{align*}
  where
  $\nabla_{\bm{y}}:=\left(\parder{y_1},\dots,\parder{y_{\dy}}\right)$.
  A useful property is that the minimizer of $J(g)$ coincides with
  $\log p(\bm{y}|\bm{x})$ up to the partition function, which is
  stated by the following proposition:
  \begin{proposition}
    \label{prop:fisher-minimizer} Assume that $p(\bm{y},\bm{x})>0$ for
    all $\bm{y}\in\calY$ ans $\bm{x}\in\calX$. Let us express the
    minimizer of $J(g)$ with respect to $g$ by
    \begin{align*}
      \widehat{g}:=\argmin_{g} J(g).
    \end{align*}   
    Then, $\widehat{g}(\bm{y},\bm{x})$ equals to $\log p(\bm{y}|\bm{x})$
    up to the partition function, i.e., $\widehat{g}(\bm{y},\bm{x})$
    satisfies the following equation:
    \begin{align}
      \nabla_{\bm{y}}\widehat{g}(\bm{y},\bm{x}) 
      =\nabla_{\bm{y}}\log q(\bm{y}|\bm{x})
    =\nabla_{\bm{y}}\log p(\bm{y}|\bm{x}).
    \label{asymptotic-solution}
  \end{align}
 \end{proposition}
 The proof can be seen in the supplementary material.

 Proposition~\ref{prop:fisher-minimizer} implies that minimising $J$
 asymptotically yields a consistent estimator. However, in practice,
 we need to estimate $J$ from finite samples. Following score
 matching~\citep{hyvarinen2005estimation} and further developments
 by~\citet{pmlr-v84-arbel18a}, we can derive an easily computable
 version of $J$ under some mild assumptions as follows:
 \begin{align*}
   \tilde{J}(g)&:=J(g)-C\\
   &=\sum_{j=1}^{\dy}\left[
     \frac{1}{2}\iint \left\{\parder{y_j}g(\bm{y},\bm{x})\right\}^2p(\bm{y},\bm{x})
     \intd\bm{y}\intd\bm{x}-\iint\left\{\parder{y_j}g(\bm{y},\bm{x})\right\}
     \parder{y_j}p(\bm{y},\bm{x})\intd\bm{y}\intd\bm{x}\right]\\
   &=\sum_{j=1}^{\dy}\left[ \frac{1}{2}\iint
     \left\{\parder{y_j}g(\bm{y},\bm{x})\right\}^2
     p(\bm{y},\bm{x})\intd\bm{y}\intd\bm{x}
     +\iint\left\{\frac{\partial^2}{\partial y_j^2}
       g(\bm{y},\bm{x})\right\}
     p(\bm{y},\bm{x})\intd\bm{y}\intd\bm{x}\right],
  \end{align*}
  where $C=\frac{1}{2}\iint\|\nabla_{\bm{y}}\log q(\bm{y}|\bm{x})\|^2
  p(\bm{y},\bm{x})\intd\bm{y}\intd\bm{x}$, and we used the relation
  $\nabla_{\bm{y}}\log q(\bm{y}|\bm{x})=\nabla_{\bm{y}}\log
  p(\bm{y}|\bm{x})$ and applied the \emph{integration by parts} on the
  second line under a mild assumption that
  $\lim_{|y_j|\rightarrow\infty}\parder{y_j}g(\bm{y},\bm{x})p(\bm{y},\bm{x})=0$
  for all $\bm{x}$ and $j$. After substituting the model
  $\bm{w}^{\top}\bm{h}$ into $g$, the empirical version of $\tilde{J}$
  can be obtained as
  \begin{align*}
   \widehat{J}(\bm{w},\bm{h})&:=\frac{1}{T}\sum_{t=1}^T\sum_{j=1}^{\dy}\left[
   \frac{1}{2}\left\{\parder{y_j}\bm{w}(\bm{y}(t))^{\top}\bm{h}(\bm{x}(t))\right\}^2
   +\frac{\partial^2}{\partial
   y_j^2}\bm{w}(\bm{y}(t))^{\top}\bm{h}(\bm{x}(t))\right].
  \end{align*}  
  Thus, an estimator can be obtained by minimising $\widehat{J}$ with
  respect to $w_i$ and $h_i$ for all $i$.
    
  \paragraph{Learning Algorithm with Neural Networks and Reproducing
    Kernels:} Motivated by
  Proposition~\ref{prop:universal-approximation}, we model $h_i(\bm{x})$
  by a feedforward neural network (fNN), $h_i(\bm{x};\bm{\theta}_i)$
  where $\bm{\theta}_i$ denotes a vector of parameters in the fNN. On
  the other hand, modelling $w_i(\bm{y})$ by fNNs is problematic because
  $\widehat{J}$ includes the partial derivatives of $w_i(\bm{y})$, and
  computation of the (particularly second-order) derivatives are highly
  complicated in learning
  fNNs~\citep{martens2012estimating}. Alternatively, following
  Proposition~\ref{prop:universal-approximation}, we estimate $w_i$ as
  an element in an RKHS with a universal kernel, which has the universal
  approximation capability~\citep{micchelli2006universal}. This approach
  greatly simplifies the derivative computation in~$\widehat{J}$ because
  of the representer theorem: Inspired by the representer theorem for
  derivatives~\citep{zhou2008derivative}\footnote{If we exactly follow
  the representer theorem in~\citet{zhou2008derivative}, $w_i$ should
  have the following form: $w_i(\bm{y}) = \sum_{t=1}^T\sum_{j=1}^{\dy}
  \alpha_{i,(j-1)T+t}
  \parder{y_j'}k(\bm{y},\bm{y}')\Bigr|_{\bm{y}'=\bm{y}(t)}
  +\beta_{i,(j-1)T+t} \frac{\partial^2}{\partial y_j^{\prime 2}}
  k(\bm{y},\bm{y}')\Bigr|_{\bm{y}'=\bm{y}(t)}$, where $\beta_{i,j}$
  denotes coefficients. To decrease the computational costs, we simplify
  the model by setting all $\beta_{i,j}=0$ and using only a randomly
  chosen subset of $\{\bm{y}(t)\}_{t=1}^T$ as the centers in the kernel
  function. The same simplification under the Fisher divergence is
  previously performed and theoretically shown not to incur the
  performance degeneration under some
  conditions~\citep{pmlr-v84-sutherland18a}.}, we employ the following
  linear-in-parameter model for $w_i$:
  \begin{align}
    w_i(\bm{y};\bm{\alpha}_i) = \sum_{b=1}^B\sum_{j=1}^{\dy} \alpha_{i,(j-1)B+b}
    \parder{y_j'}k(\bm{y},\bm{y}')\Bigr|_{\bm{y}'=\tilde{\bm{y}}(b)}
    =\bm{\alpha}_i^{\top}\bm{k}'(\bm{y}),
   \label{kernel-model}
  \end{align}
  where $k$ and $\alpha_{i,j}$ are the kernel function and
  coefficients respectively. To decrease the computation cost, we
  follow the Nyst{\"o}m approximation~\citep{rudi2015less} and use
  $\{\tilde{\bm{y}}(b)\}_{b=1}^B$ as the center points in $k$ by
  randomly choosing $B (\ll T)$ points from $\{\bm{y}(t)\}_{t=1}^T$.
    
  By substituting an fNN $h_i(\bm{x};\bm{\theta}_i)$ and the
  linear-in-parameter model $w_i(\bm{y};\bm{\alpha}_i)$ into
  $\widehat{J}(\{w_i, h_i\}_{i=1}^d)$, the optimal parameters are
  obtained as
  \begin{align*}
    \{\widehat{\bm{\theta}}_i, \widehat{\bm{\alpha}}_i\}_{i=1}^d:=
    \argmin_{\{\bm{\theta}_i,\bm{\alpha}_i\}_{i=1}^d}
    \widehat{J}(\{\bm{\theta}_i,\bm{\alpha}_i\}_{i=1}^d).
  \end{align*}
  Setting
  $\widehat{\bm{w}}(\bm{y}):=(w_1(\bm{y};\widehat{\bm{\alpha}}_1),\dots,w_d(\bm{y};\widehat{\bm{\alpha}}_d))^{\top}$
  and
  $\widehat{\bm{h}}(\bm{x}):=(h_1(\bm{x};\widehat{\bm{\theta}}_1),\dots,h_d(\bm{x};\widehat{\bm{\theta}}_d))^{\top}$,
  the estimator is finally given by
  \begin{align*}
    \widehat{\bm{w}}(\bm{y})^{\top}\widehat{\bm{h}}(\bm{x})
    =\sum_{i=1}^d w_i(\bm{y};\widehat{\bm{\alpha}}_i)
    h_i(\bm{x};\widehat{\bm{\theta}}_i).
  \end{align*}
  We call this method the \emph{neural-kernelized conditional density
    estimator} (NKC).
 \section{Simulations}
 \label{sec:simulations}
 This section investigates the practical performance of NKC, and
 compares it with existing methods.
 \vspace{-2mm}
 \paragraph{Algorithms, settings, and evaluation:} We report the
 results on the following three methods:~~~~~~~~~~~~~~
 \begin{itemize}\setlength{\leftskip}{-0.8cm}
 \item {\bf NKC (Proposed method)}: $\bm{h}$ was modelled by a
   feedforward neural network with three layers: The numbers of hidden
   units in the three layers were $100, 50$ and $d$, where $d$ was
   either $3$ or $5$ (we performed preliminary experiments for $d=1$
   as well, but the results were often worse than $d=3, 5$). The
   activation functions were all ReLU. For $\bm{w}$, we employed the
   Gaussian kernel, and fixed the number of the kernel centers at
   $B=100$. To perform stochastic gradient descent,
   RMSprop~\citep{Hinton2012} was used with $128$ minibatches. When
   learning $\bm{h}$ and $\bm{w}$, model selection was also performed
   for the learning rate and the width parameter in the Gaussian
   kernel with a early stopping technique: The datasets were first
   divided into the training (80\%) and validation (20\%) datasets,
   and then $\bm{h}$ and $\bm{w}$ were learned using the training
   datasets for $100$ epochs. Finally, the best epoch result was
   chosen as the final result using the validation dataset in terms of
   $\widehat{J}$. Other details are given in the supplementary
   material.
   
 \item {\bf Conditional variational autoencoder
     (CVAE)~\citep{sohn2015learning}:} This is a conditional density
   estimator based on variational methods, shown for comparison. In
   CVAE, three conditional densities related to latent variables,
   which are called recognition, prior and generation networks, were
   modelled by parametric Gaussian densities: the means and the
   logarithms of standard deviations in the Gaussian densities of the
   recognition and generation networks were expressed by neural
   networks, while the prior network was modelled by the normal
   density independent of $\bm{x}$ as
   in~\citep{kingma2014semi,sohn2015learning}. The neural networks had
   two layers with $50$ hidden units. The activation function was ReLU
   in the middle layer, while no activation function was applied in
   the final layer. The dimensionality of latent variables was
   $50$. The same optimisation procedures as in NKC was applied. It
   should be noted that the number of parameters in CVAE was a bit,
   but not much, larger than in NKC (the exact numbers of parameters
   are given in the supplementary material).

 \item {\bf
     LSCDE}~\citep{sugiyama2010least}\footnote{\url{http://www.ms.k.u-tokyo.ac.jp/software.html\#LSCDE}:}
   This is a least-squares estimation method for conditional densities
   based on the Gaussian kernel, shown for comparison.  In the default
   setting, LSCDE only uses $\min(100, T)$ centers in the Gaussian
   kernel~\citep{sugiyama2010least}, but to apply LSCDE to a large
   amount of and high-dimensional data, we set $\min(1000, T)$
   centers, which were randomly chosen from the whole data. The
   bandwidth in the Gaussian kernel, as well as the regularisation
   parameters, were determined by five-fold cross-validation.
 \end{itemize}
 The performance of each method was evaluated by the log-likelihood
 using test data. To estimate the log-likelihood in NKC, we estimated
 the partition function by importance sampling, while the
 log-likelihood in CVAE is estimated as in~\citet[Eq.(6)]{sohn2015learning}.
 
\vspace{-2mm}
\paragraph{Multimodal Density Estimation on Artificial Data:}
Conditional density estimation is particularly useful compared to
simple conditional mean estimation in the case where conditional
densities are multimodal. Therefore, we first numerically demonstrate
the performance of NKC on multimodal density estimation. To this end,
we first generated data samples of $\bm{x}$ from the normal density,
and then data samples of $y$ was drawn from a mixture of three
Gaussians: $p(y|\bm{x})=\sum_{k=1}^3 \frac{c_k}{\sqrt{2\pi\sigma^2}}
\exp\left(-\frac{(y-\mu_k(\bm{x}))^2}{2\sigma^2}\right)$ where
$\dy=1$, $c_k=1/3$ for all $k$ and $\sigma=0.25$. The dimension $\dx$
took the value $50$, $100$, $200$ or $300$. Here, $\mu_k(\bm{x})$ were
modelled by a three layer neural network with leaky ReLU where the
number of hidden units is $\dx$, $\dx$ and $1$, and the parameter
values were randomly determined in each run. The total number of
samples was fixed at $T=30,000$.

The results are presented in the left panel of Table~\ref{illust-CDE}.
As $\dx$ increases, the performance of LSCDE gets worse, as would be
expected from a method based on Gaussian kernels. On the other hand,
NKC and CVAE work well even in $\dx=100$. When $\dx=200, 300$, NKC
performs better than CVAE.  Thus, NKC should be a useful method for
multimodal density estimation with high-dimensional data. Regarding
the computational time, LSCDE was the fastest because the optimal
solution in LSCDE is analytically computed. On the other hand, NKC and
CVAE had comparable time if any model selection was not used.
 
 \begin{table}
   \caption{\label{illust-CDE} Simulation on artificial data (left panel)
     and experiments on benchmark datasets (right panel).  The mean values of the
     log-likelihood over $10$ runs. A larger value indicates a better
     result.  The best and comparable methods judged by the Wilcoxon
     signed-rank test at the significance level 5\% are described in
     boldface. NKC$_{3}$ and NKC$_{5}$ denote NKC with $d=3$ and $d=5$,
     respectively.}
   \begin{minipage}[t]{0.52\textwidth}
     \begin{tabular}{|l|c|c|c|c|}
       & NKC$_3$ & NKC$_5$ & CVAE & LSCDE \\ \hline 
       $\dx=50$ & {\bf -0.75} & {\bf -0.75} & {\bf -0.79} & -0.96 \\ 
       $\dx=100$ & {\bf -0.76} & {\bf -0.76} & {\bf -0.77} & -1.25 \\ 
       $\dx=200$ & {\bf -0.81} & {\bf -0.82} & -0.89 & -1.28 \\ 
       $\dx=300$ & {\bf -0.81} & {\bf -0.82} & -0.93 & -1.69 
     \end{tabular}
   \end{minipage}
   \begin{minipage}[t]{0.48\textwidth}
     \begin{center}
       \begin{tabular}{|l|c|c|c|c|}
         & NKC$_{3}$ & NKC$_{5}$ & CVAE & LSCDE \\ \hline 
         Ail. &  {\bf -0.52} & {\bf -0.52} & {\bf -0.50} & -1.00 \\  
         Fri. &  {\bf -0.34} & {\bf -0.34} & -0.41 & -1.01 \\  
         Kin. & {\bf -0.62} & {\bf -0.63} & -0.65 & -0.96 \\ 
         Pum. &  -1.53 & -1.58 & {\bf -0.97} & -1.59 
       \end{tabular} 
     \end{center}    
   \end{minipage}
 \end{table} 
 \vspace{-2mm}
 \paragraph{Benchmark Dataset:}
 Second, we evaluate the performance of NKC on benchmark datasets. The
 datasets were the publicly available\footnote{\url{http://www.dcc.fc.up.pt/~ltorgo/Regression/DataSets.html}} datasets
 Ailerons $(\dx=26, T=13,750)$, Friedman $(\dx=10, T=40,768)$,
 Kinematics$(\dx=8, T=8,192)$, and Pumadyn$(\dx=32, T=8,192)$. Before
 performing conditional density estimation, we first standardised the
 datasets by subtracting the means and dividing each variable by its standard
 deviation. Then, in each dataset, 10\%
 of the data samples was used for test and the remaining
 data samples were for training. 

 The results are summarised in the right panel of
 Table~\ref{illust-CDE}, showing that NKC compares favourably with
 CVAE, and clearly beats LSCDE.
 \section{Insights on Representation Learning}
  Next we provide interesting insights on how NKC is related to
  representation learning. In fact, we show that NKC has interesting
  connections to two frameworks called \emph{sufficient dimension
    reduction} (SDR) and \emph{independent component analysis}
  (ICA). More specifically, we prove that the internal representation
  given by $\widehat{\bm{h}}$ in NKC can provide a useful
  lower-dimensional representation of $\bm{x}$ for supervised
  learning, if the dimension of the internal representation is
  restricted. On the other hand, if the data comes from a nonlinear
  ICA model, and the dimension is not reduced, the internal
  representation corresponds to an estimate of nonlinear independent
  components in unsupervised learning.
  \subsection{Nonlinear Sufficient Dimension Reduction}
  \paragraph{Problem Formulation in Sufficient Dimension Reduction:}
  \emph{Sufficient dimension reduction} (SDR) is a rigorous framework
  of \emph{supervised} dimension
  reduction~\citep{li1991sliced,cook1998regression}. The goal of SDR
  is to estimate a lower-dimensional representation of $\bm{x}$
  satisfying the following condition:
  \begin{align}
    p(\bm{y}|\bm{x})=p(\bm{y}|\bm{z})\quad\text{or equivalently}\quad
    \bm{y}\perp\bm{x}|\bm{z}, \label{SDRcond}
  \end{align}
  where $\perp$ denotes statistical independence,
  $\bm{z}=(z_1,\dots,z_{d})^{\top}$, and $d<\dx$. The SDR
  condition~\eqref{SDRcond} intuitively means that $\bm{z}$ has the
  same amount of information to predict $\bm{y}$ as $\bm{x}$.
  \paragraph{Connection to SDR:}
  Here, we show a connection of NKC to nonlinear SDR. In fact, the
  following theorem, which is proven in the supplementary material,
  implies that $\widehat{\bm{h}}$ in NKC is a useful lower-dimensional
  representation on SDR:
  \begin{theorem}
    \label{theo:SDR} Suppose that $d<\dx$, there exists a
    lower-dimensional representation $\bm{z}$ such that the SDR
    condition~\eqref{SDRcond} is satisfied, the domains of $\bm{y}$
    and $\bm{x}$ are compact, $\log q(\bm{y}|\bm{x})$ and
    $\bm{w}(\bm{y})^{\top}\bm{h}(\bm{x})$ are both continuous, and in
    the limit of infinite data, $\nabla_{\bm{y}}\log q(\bm{y}|\bm{x})$
    is universally approximated as
    \begin{align}
     \nabla_{\bm{y}} \log q(\bm{y}|\bm{x})
     =\nabla_{\bm{y}}\bm{w}(\bm{y})^{\top}\bm{h}(\bm{x}) =\sum_{i=1}^d
     \nabla_{\bm{y}}w_i(\bm{y})h_i(\bm{x}).
    \label{universal-approximator}
   \end{align}
    Then, $\bm{h}$ satisfies the SDR condition.
  \end{theorem}
  
  Propositions~\ref{prop:universal-approximation}
  and~\ref{prop:fisher-minimizer} imply that
  $\nabla_{\bm{y}}\widehat{\bm{w}}(\bm{y})^{\top}
  \widehat{\bm{h}}(\bm{x})$ approximates $\nabla_{\bm{y}}\log
  q(\bm{y}|\bm{x})$ well as the number of data samples
  increases. Thus, the universal approximation
  assumption~\eqref{universal-approximator} would be
  realistic. Interestingly, Thereom~\ref{theo:SDR} shows that NKC
  estimates the log-conditional density, while implicitly performing
  nonlinear dimensionality reduction. This would imply that if there
  exists a lower-dimensional structure (manifold), the log-conditional
  density can be accurately estimated with a small $d$, i.e., small
  parameters. To the best of our knowledge, this work is the first
  attempt to explicitly connect neural networks to nonlinear SDR.
  \paragraph{Relation with Existing Nonlinear SDR Methods:}
  \label{ssec:relation-SDR}
  \emph{Kernel sliced inverse regression} (KSIR) is a nonlinear
  extension of a classical linear SDR method called \emph{sliced
    inverse regression} (SIR)~\citep{li1991sliced} via the
  \emph{kernel
    trick}~\citep{wu2008kernel,yeh2009nonlinear,wu2007regularized}. As
  in kernel PCA~\citep{scholkopf2002learning}, input data $\bm{x}$ is
  mapped to a high-dimensional feature space $\mathcal{F}$ by
  $\bm{\phi}(\bm{x})$, and then SIR is performed in $\mathcal{F}$. As
  in SIR, KSIR makes an assumption that the probability density
  function of $\bm{\phi}(\bm{x})$ is elliptically
  symmetric~\citep[pp.594]{wu2008kernel}, while NKC does not have
  explicit assumptions on densities. Other nonlinear SDR methods have
  been proposed in~\citet{lee2013general}. A drawback of these methods
  as well as KSIR require to compute the eigenvectors of a $T$ by $T$
  matrix. Therefore, it is not straightforward to apply them to a
  large amount of data. In contrast, a possible advantage of NKC would
  be applicability to large datasets.  
  \subsection{Nonlinear Independent Component Analysis}
  \paragraph{Problem Formulation and Background in Independent
    Component Analysis:}
  Independent component analysis (ICA) is a successful framework in
  unsupervised learning. ICA assumes that data $\bm{x}$ is generated from
  \begin{align}
   \bm{x}=\bm{f}(\bm{s}), \label{ICA-model}
  \end{align}
  where $\bm{f}(\bm{s})=(f_1(\bm{s}), \dots, f_d(\bm{s}))^{\top}$,
  $f_i$ denotes a smooth and invertible nonlinear function, and
  $\bm{s}=(s_1,\dots, s_d)^{\top}$ is a vector of the source
  components. In classical ICA, $s_i$ are assumed to be statistically
  independent each other. Then, the problem is to estimate $\bm{f}$ or
  $\bm{s}$ from observations of $\bm{x}$ only. When
  $\bm{f}(\bm{s})=\bm{A}\bm{s}$ with an invertible $d$ by $d$ matrix
  $\bm{A}$, the identifiability conditions are
  well-understood~\citep{comon1994independent}.  On the other hand,
  for general nonlinear functions $f_i$, the problem is considerably
  more difficult: There exist infinitely many solutions and therefore
  the problem is ill-posed without any additional
  information~\citep{hyvarinen1999nonlinear}. Recently, some
  identifiability conditions were found
  by~\citet{hyvarinen2016unsupervised,pmlr-v54-hyvarinen17a}. The key
  idea is to use time information as additional information. Below, we
  show novel identifiability conditions by connecting NKC with
  nonlinear ICA.
  \paragraph{Connection to Nonlinear ICA:}
  Here, suppose that $\bm{y}$ contains some additional variables which
  are not of primary interest, but are dependent on $\bm{x}$, thus
  providing extra information that helps in identifiability. For
  example, following \citet{hyvarinen2016unsupervised}, $\bm{y}$ could
  be the time index for time series data.  We show next that
  $\widehat{\bm{h}}$ in NKC recovers the source vector $\bm{s}$ up to
  some indeterminancies with a new identifiability proof. To this end,
  we make the following assumptions:
  \begin{enumerate}[(I1)]
  \item[(I1)] Given $\bm{y}$, $s_i$ are statistically independent each
	      other: $p(\bm{s}|\bm{y})=\prod_{i=1}^{d} p(s_i|\bm{y})$.

  \item[(I2)] The conditional density of $s_i$ given $\bm{y}$ belongs to
    the exponential family as
    \begin{align*}
      \log p(s_i|\bm{y})=\lambda_{i}(\bm{y}) \tilde{q}_{i}(s_i)-\log
      Z(\lambda_{i}(\bm{y})),
    \end{align*}
    where $Z$ is the partition function and $\tilde{q}_{i}$ is a
    nonlinear scalar function.
    
  \item[(I3)] There exists the inverse of the generative model, i.e.,
    $\bm{s}=\bm{g}(\bm{x})$ where $\bm{g}:=\bm{f}^{-1}$.
    
  \item[(I4)] In the limit of infinite data, $\nabla_{\bm{y}}\log
    q(\bm{y}|\bm{x})$ is universally approximated as
    in~\eqref{universal-approximator}.

  \item[(I5)] There exist $N$ points, $\bm{y}(1),
    \bm{y}(2),\dots,\bm{y}(N)$ such that the following $d$ by $d$
    matrix is invertible:
    $\sum_{n=1}^N\nabla_{\bm{y}}\bm{\lambda}(\bm{y}(n))
    (\nabla_{\bm{y}}\bm{\lambda}(\bm{y}(n)))^{\top}$, where
    $\bm{\lambda}(\bm{y}):=(\lambda_{1}(\bm{y}),\dots,
    \lambda_{d}(\bm{y}))^{\top}$.
  \end{enumerate}
  Then, the following theorem elucidates the relationship between
  NKC and nonlinear ICA:
  \begin{theorem}
    \label{theo:ICA} Suppose that assumptions (I1-I5) hold. Then, in
    the limit of infinite data, $\bm{h}(\bm{x})$
    in~\eqref{universal-approximator} equals to $\bm{q}(\bm{s})$ up to
    an invertible linear transformation, i.e.,
   \begin{align*}
     \tilde{\bm{q}}(\bm{s})=\bm{B}\bm{h}(\bm{x})+\bm{b},
   \end{align*}   
   where
   $\tilde{\bm{q}}(\bm{s})=(\tilde{q}_{1}(s_1),\tilde{q}_{2}(s_2),\dots,\tilde{q}_{d}(s_{d}))^{\top}$,
   $\bm{B}\in\R{d\times d}$ and $\bm{b}\in\R{d}$.
  \end{theorem}
  The proof can be seen in the supplementary material:

  \paragraph{Relation with Existing Nonlinear ICA Methods:}
  Theorem~\ref{theo:ICA} states that $\tilde{q}_i(s_i)$ are identified
  up to linear transformation, as in the previous proof for a nonlinear
  ICA method called \emph{time contrastive learning}
  (TCL)~\citep{hyvarinen2016unsupervised}. Thus, in practice, after
  estimating $\log q(\bm{y}|\bm{x})$, we might apply some linear ICA
  method to $\widehat{\bm{h}}(\bm{x})$ to get closer to the
  independent components. For that purpose we should further assume
  the $s_i$ are marginally independent, while the Theorem assumed only
  their conditional independent. The assumption of an exponential
  family model for the sources is also very similar to TCL. The
  fundamental difference of the present Theorem compared to TCL is
  that here we have a continuous-valued additional variable $\bm{y}$,
  while TCL is based on a discrete variable (time-segmentation
  label). Thus, the current proof is related to the TCL theory but not
  a simple modification.
  \section{Conclusion}
  We proposed a novel method based on score matching to estimate
  conditional densities in a general setting where no particular
  (parametric) assumptions on the density need to be made. We
  developed a novel neural-kernelized approach, which combines neural
  networks with reproducing kernels to apply score matching. In
  particular, using neural networks allows for powerful function
  approximation which is scalable for large dataset, while using
  kernels avoids the complications related to the derivatives needed
  in score matching.  We proved that the model combining neural
  networks with reproducing kernels has the universal approximation
  capability, and that the ensuing estimation method is
  consistent. The practical performance of the proposed method was
  investigated both on artificial and benchmark datasets, and it was
  useful in high-dimensional conditional density estimation and
  compared favourably to existing methods. Finally, we showed that the
  proposed method has interesting connections to two rigorous
  probabilistic frameworks of representation learning, nonlinear
  sufficient dimension reduction and nonlinear independent component
  analysis, thus opening new avenues for theoretical analysis of
  representation learning.
 \bibliography{../../../papers}
 \bibliographystyle{abbrvnat}
 \newpage
 \appendix
 \restylefloat{table}
 \section{Proof of Proposition~\ref{prop:fisher-minimizer}} 
 \begin{proof}
  The minimizer $\widehat{g}$ should satisfy the following optimality
  condition for the G{\^a}teaux derivative of $J$ along with an
  arbitrary direction $f$:
  \begin{align}
   \frac{\intd}{\intd\epsilon}J(\widehat{g}+\epsilon
   f)\Bigr|_{\epsilon=0}=0.  \label{Gateaux-derivative}
  \end{align}
  
  The left-hand side of~\eqref{Gateaux-derivative} shows that
  \begin{align*}
   &J(\widehat{g}+\epsilon f)\\
   &=J(\widehat{g})
   +\epsilon\sum_{j=1}^{\dy}\int_{\calY}\int_{\calX}
   \left\{\parder{y_j}f(\bm{y},\bm{x})\right\}
   \left\{\parder{y_j}\widehat{g}(\bm{y},\bm{x})
     -\parder{y_j}\log q(\bm{y}|\bm{x})\right\}
   p(\bm{y},\bm{x}) \intd\bm{y}\intd\bm{x}+o(\epsilon^2).
 \end{align*}
 Since $f$ is arbitrary and $p(\bm{y},\bm{x})>0$, substituting the
 equation above into the optimality
 condition~\eqref{Gateaux-derivative} yields
 \begin{align}
   \parder{y_j}\widehat{g}(\bm{y},\bm{x})
   =\parder{y_j}\log q(\bm{y}|\bm{x})
   =\parder{y_j}\log p(\bm{y}|\bm{x}), \quad\forall j.   
   \label{sufficient-condition}
  \end{align}  
  Furthermore, the positive assumption of $p(\bm{y},\bm{x})$ ensures
  that $\widehat{g}$ is the unique minimizer. Thus, the proof is
  completed.
 \end{proof}
 \section{Simulation Details}
 Regarding simulations on artificial data, in model selection, we set
 the candidates of the learning rate and width parameter at
 $\{10^{-3}, 5\times10^{-4}, 10^{-4}, 5\times10^{-5}, 10^{-5} \}$ and
 $\{0.5, 1.0, 3.0, 5.0\}$, respectively. The $\ell_2$ regularisation
 was applied to only $\{\bm{\alpha}_i\}_{i=1}^d$ in the kernel part
 $\bm{w}$ with $10^{-4}$ decay, while no regularisation was used in
 the neural network part $\bm{h}$. For benchmark datasets, the
 candidates of the learning rate is the same , but the width parameter
 is selected from $\{0.5, 1.0, 2.0, 3.0\}$ multiplied by
 $\sigma_{\mathrm{med}}^{\mathrm{y}}$ which is the median of
 $|y(t)-y(t')|$ with respect to $t$ and $t'$. In addition, no $\ell_2$
 regularisation was used for any parameters in benchmark datasets.

 The following table shows the number of parameters of NKC and CVAE in
 simulations on artificial data (Section~\ref{sec:simulations}).
 \begin{table}[H]
   \begin{center}
     \begin{tabular}{c|c|c|c}
       & NKC$_{3}$ & NKC$_{5}$ & CVAE  \\ \hline 
       $\dx=50$ & $10603$ & $10905$ & $12852$  \\ 
       $\dx=100$ & $15603$ & $15905$ & $17852$ \\ 
       $\dx=200$ & $25603$ & $25905$ & $27852$ \\ 
       $\dx=300$ & $35603$ & $35905$ & $37852$ \\ 
     \end{tabular} 
   \end{center}    
 \end{table}
 \section{Proof of Theorem~\ref{theo:SDR}}
 \label{sec:proof-SDR}
 \begin{proof}   
   We first use the standard path integral
   formula~\citep{strang1991calculus}: For the vector field
   $\nabla_{\bm{y}} \log q(\bm{y}|\bm{x})$ and a differentiable curve
   ${\bm\gamma}(t),\,t\in[0,s]$ connecting a fixed point
   $\bm{y}_{\mathrm{c}}$ and arbitrary point $\bm{y}$,
   \begin{align*}
     \int_0^s \<\nabla_{\bm{y}} \log
     q({\bm\gamma}(t)|\bm{x}),\,\dot{\bm\gamma}(t)\> \intd t
     =\log q(\bm{y}|\bm{x})- \log q(\bm{y}_{\mathrm{c}}|\bm{x}),
   \end{align*}
   where $\dot{\bm\gamma}(t)=\frac{\intd}{\intd t}\bm\gamma(t)$ and
   $\<\cdot,\cdot\>$ denotes the inner product. Then, applying the
   path integral formula to both sides in the universal approximation
   assumption~\eqref{universal-approximator} yields
   \begin{align*}
     \log q(\bm{y}|\bm{x}) = \bm{w}(\bm{y})^{\top}\bm{h}(\bm{x})
     +A(\bm{x}),
   \end{align*}
   where $A(\bm{x}):=\log
   q(\bm{y}_{\mathrm{c}}|\bm{x})-\bm{w}(\bm{y}_{\mathrm{c}})^{\top}\bm{h}(\bm{x})$,
   and is independent of $\bm{y}$ because $\bm{y}_{\mathrm{c}}$ is a
   fixed point. This equation gives
   \begin{align}
     q(\bm{y}|\bm{x})\cdot\exp\left(-A(\bm{x})\right)
     =\exp\left(\bm{w}(\bm{y})^{\top}\bm{h}(\bm{x})\right).   
   \end{align}
   Integrating the both sides yields
  \begin{align*}
    Z_{\mathrm{x}}(\bm{x}):=\int
    q(\bm{y}|\bm{x})\cdot\exp\left(-A(\bm{x})\right) \intd\bm{y}= \int
    \exp\left(\bm{w}(\bm{y})^{\top}\bm{h}(\bm{x})\right)\intd\bm{y}
    =:Z_{\mathrm{h}}(\bm{h}).
  \end{align*}
  
  By the definition of the conditional density,
  \begin{align*}
   p(\bm{y}|\bm{x}) = 
   \frac{q(\bm{y}|\bm{x})\cdot\exp\left(-A(\bm{x})\right)}
   {Z_{\mathrm{x}}(\bm{x})}
   =\frac{\exp\left(\bm{w}(\bm{y})^{\top}\bm{h}(\bm{x})\right)}
   {Z_{\mathrm{x}}(\bm{x})}
   =\frac{\exp\left(\bm{w}(\bm{y})^{\top}\bm{h}(\bm{x})\right)}
   {Z_{\mathrm{h}}(\bm{h})}
   =p(\bm{y}|\bm{h})
  \end{align*}
  Thus, $\bm{h}$ satisfies the SDR condition~\eqref{SDRcond}, and the
  proof is completed.  
 \end{proof}
 \section{Proof of Thereom~\ref{theo:ICA}}
 \begin{proof}
  Bayes theorem shows that
  \begin{align*}
    \log p(\bm{x}|\bm{y}) = \log p(\bm{y}|\bm{x})+\log p(\bm{x})-\log
    p(\bm{y}).
  \end{align*}
  After taking the gradient with respect to $\bm{y}$ above, the
  universal approximator assumption (I4) gives
  \begin{align}
   \nabla_{\bm{y}}\log p(\bm{x}|\bm{y})&= \nabla_{\bm{y}}\log
   q(\bm{y}|\bm{x})-\nabla_{\bm{y}}\log p(\bm{y})
   =(\nabla_{\bm{y}}\bm{w}(\bm{y}))^{\top}
   \bm{h}(\bm{x})-\nabla_{\bm{y}}\log p(\bm{y}),\label{pdf-x-given-y1}
  \end{align}
  where the $(i,j)$-th element in $\nabla_{\bm{y}} \bm{w}(\bm{y})$ is
  $\parder{y_j}w_i(\bm{y})$.  Interestingly, the gradient with respect
  to $\bm{y}$ deletes $\log p(\bm{x})$ in~\eqref{pdf-x-given-y1}. On
  the other hand, the exponential family assumption yields another
  form of $\log p(\bm{x}|\bm{y})$ as
  \begin{align*}
   \log p(\bm{x}|\bm{y})=\sum_{i=1}^{d}\left[
   \lambda_{i}(\bm{y})\tilde{q}_{i}(g_i(\bm{x}))
   +\log|\bm{J}{\bm{g}}(\bm{x})|-\log Z(\lambda_{i}(\bm{y}))\right], 
  \end{align*}
  where $\bm{Jg}(\bm{x})$ denotes the Jacobian. By taking the gradient
  with respect to $\bm{y}$, we have
  \begin{align}
    \nabla_{\bm{y}}\log p(\bm{x}|\bm{y})=
    \sum_{i=1}^{d}\left[\nabla_{\bm{y}}\lambda_{i}(\bm{y})
      \tilde{q}_{i}(g_i(\bm{x})) -\nabla_{\bm{y}}\log
      Z(\lambda_{i}(\bm{y}))\right]. \label{pdf-x-given-y2}
  \end{align}
  A notable point is that $\log|\bm{J}{\bm{g}}(\bm{x})|$ disappeared
  after taking the gradient. Equating \eqref{pdf-x-given-y1} with
  \eqref{pdf-x-given-y2} yields
  \begin{align}
    (\nabla_{\bm{y}} \bm{\lambda}(\bm{y}))^{\top}
    \tilde{\bm{q}}(\bm{s}) -\sum_{i=1}^{d} \left\{\nabla_{\bm{y}} \log
      Z(\lambda_{i}(\bm{y}))\right\}
    =(\nabla_{\bm{y}}\bm{w}(\bm{y}))^{\top}\bm{h}(\bm{x})
    -\nabla_{\bm{y}}\log p(\bm{y}), \label{equation}
  \end{align}      
  where the $(i,j)$-th element in $\nabla_{\bm{y}}
  \bm{\lambda}(\bm{y})$is $\parder{y_j}\lambda_i(\bm{y})$.  By taking
  the summation over $N$ points $\bm{y}(1), \bm{y}(2),\dots,
  \bm{y}(N)$ after multiplying $\nabla_{\bm{y}}\bm{\lambda}(\bm{y})$
  to the both sides of~\eqref{equation}, we have
  \begin{align*}
   \left[\sum_{n=1}^N\nabla_{\bm{y}}\bm{\lambda}(\bm{y}(n))
   (\nabla_{\bm{y}} \bm{\lambda}(\bm{y}(n)))^{\top}
   \right]&\tilde{\bm{q}}(\bm{s})
   = \left[\sum_{n=1}^N\nabla_{\bm{y}}\bm{\lambda}(\bm{y}(n))
   (\nabla_{\bm{y}}\bm{w}(\bm{y}(n)))^{\top}\right]\bm{h}(\bm{x})\\
   &\hspace{-10mm}+\sum_{n=1}^N\nabla_{\bm{y}}\bm{\lambda}(\bm{y}(n))
   \left[\sum_{i=1}^{d} \left\{\nabla_{\bm{y}} \log
   Z(\lambda_{i}(\bm{y}(n)))\right\}-\nabla_{\bm{y}}\log p(\bm{y}(n))\right].
  \end{align*}
  Taking the inverse of
  $\bm{E}:=\left[\sum_{n=1}^N\nabla_{\bm{y}}\bm{\lambda}(\bm{y}(n))
    (\nabla_{\bm{y}} \bm{\lambda}(\bm{y}(n)))^{\top}\right]$ on both
  sides completes the proof as
  \begin{align*}
    \tilde{\bm{q}}(\bm{s})&= \underbrace{\bm{E}^{-1}
      \left[\sum_{n=1}^N\nabla_{\bm{y}}\bm{\lambda}(\bm{y}(n))
        (\nabla_{\bm{y}}\bm{w}(\bm{y}(n)))^{\top}\right]}_{\bm{B}}\bm{h}(\bm{x})\\
    &+\underbrace{\bm{E}^{-1}\sum_{n=1}^N\nabla_{\bm{y}}\bm{\lambda}(\bm{y}(n))
      \left[\sum_{i=1}^{d} \left\{\nabla_{\bm{y}} \log
          Z(\lambda_{i}(\bm{y}(n)))\right\}-\nabla_{\bm{y}}\log
        p(\bm{y}(n))\right].  }_{\bm{b}}
  \end{align*}
 \end{proof}
\end{document}